\documentclass[sensors,article,accept,pdftex,moreauthors]{Definitions/mdpi}

\usepackage{amsmath}
\allowdisplaybreaks

\firstpage{1} 
\pubvolume{1}
\issuenum{1}
\articlenumber{0}
\pubyear{2024}
\copyrightyear{2024}
 \externaleditor{Academic Editor:  }
\datereceived{ } 
\daterevised{ } 
\dateaccepted{ } 
\datepublished{ } 
\hreflink{https://doi.org/} 

\Title{Multi-AUV Kinematic Task Assignment Based on Self-Organizing Map Neural Network and Dubins Path~Generator}

\TitleCitation{Multi-AUV Kinematic Task Assignment Based on Self-Organizing Map Neural Network and Dubins Path Generator}

\Author{{Xin}
 Li $^{1,}$*\orcidA{}, Wenyang Gan $^{2}$\orcidC{}, Pang Wen $^{3}$ and Daqi Zhu $^{3}$\orcidB{}}

\AuthorNames{Xin Li, Wenyang Gan, Wen Pang and Daqi Zhu }

\AuthorCitation{{Xin,} 
 L.; Wenyang, G.; Wen, P.; Daqi, Z.}

\address{%
$^{1}$ \quad College of Information Engineering, Shanghai Maritime University, 1550 Haigang Avenue, Pudong New Area, 
 Shanghai 201306, China\\
$^{2}$ \quad Logistics Engineering College, Shanghai Maritime University, Shanghai 201306, China; {wygan@shmtu.edu.cn} 
\\
$^{3}$ \quad Mechanics Institute, University of Shanghai for Science and Technology, Shanghai 200093, China; {wenpang@ieee.org (P.W.); zdq367@aliyun.com (D.Z.)}}

\corres{Correspondence: xinli@shmtu.edu.cn} 

\abstract{To deal with the task assignment problem of multi$-$AUV systems under kinematic constraints, which means steering capability constraints for underactuated AUVs or other vehicles likely, an improved task assignment algorithm is proposed combining the Dubins Path algorithm with improved SOM neural network algorithm.
At first, the aimed tasks are assigned to the AUVs by the improved SOM neural network method based on workload balance and neighborhood function.
When there exists kinematic constraints or obstacles which may cause failure of trajectory planning, task re$-$assignment will be implemented by changing the weights of SOM neurals, until the AUVs can have paths to reach all the targets.
Then, the Dubins paths are generated in several limited cases. The AUV's yaw angle is limited, which results in new assignments to the targets. Computation flow is designed so that the algorithm in MATLAB and Python 
 can realize the path planning to multiple targets. Finally, simulation results prove that the proposed algorithm can effectively accomplish the task assignment task for a multi$-$AUV system.}


\begin{document}


\section{Introduction}

At the present time, there have been a lot of theoretical and practical research results on the task assignment algorithms for the multi$-$AUV system.
The task assignment algorithm covers two aspects: 1. task allocation and 2. path planning, which refers to the technology of controlling a group of AUVs according to a certain algorithm and reaching the target along the optimized path under the kinematic constraints.
In other words, task allocation ensures optimized task assignment such as the Traveling Salesman Problem (TSP); path planning guarantees the feasibility for the agents moving to the assigned tasks.
For underwater vehicles, task assignment refers to the assignment of a number of targets to a multi$-$AUV system, and~the cost of the whole multi$-$AUV system is the minimum under the premise that all targets are traversed.
The path planning technique refers to the achievement method for any single AUV in the group after global task assignment~\cite{WOS:001200174800001}.

The market mechanism algorithm is one the most common methods to solve the multitask assignment operations~\cite{2015Market}.
Another commonly used task assignment algorithm is the ant colony algorithm which mainly studies the autonomous task assignment of multi$-$robot systems in dynamic environments~\cite{2013The}. 
These are also other task assignment methods, such as that with intelligent heuristic algorithm handling water flow influence~\cite{Zhu_2024}, while kinematic path planning is not considered for most of them. As~for neural networks, study~\cite{2011Multi} further proposed a task allocation algorithm suitable for large$-$scale multi$-$robot clusters. The~self$-$organizing neural network algorithm can also be applied to multi$-$task assignment. The~SOM algorithm was proposed by a Finnish scholar, Teuvo Kohonen, in~the 1980s~\cite{1982Analysis,2020Particle}.
References~\cite{1687936,SOM2012} applied the self$-$organizing neural network to the task assignment and distribution of multi$-$mobile robot systems. Using the SOM neural network algorithm, dynamic task assignments for multi$-$AUV systems in two$-$dimensional ocean environments are implemented. Then, the~multi$-$AUV multi$-$target allocation strategy of self$-$organizing map (SOM) neural network is further extended and applied to the three$-$dimensional marine environment~\cite{8294296}.  {{Event$-$triggered} 
 adaptive neural network tracking control for uncertain systems was proposed in~\cite{liu_event-triggered_2024}, where the event$-$triggering mechanism can reduce the update rate of input signals and avoid the Zeno behavior.}

Path planning is one of the intelligent behaviors that AUVs need to possess. The~so$-$called path planning means that in an environment with obstacles, the~AUV finds a collision$-$free path from the initial state to the target state according to certain evaluation criteria~\mbox{\cite{1999A,2015Online}}.
Traditional path planning methods mainly include the visual graph method, artificial potential field method, genetic algorithm, fuzzy logic method, etc.~\cite{1986Real,1998GA,2000Motion}.
These path planning methods have their own scope of application. However, underwater movements of the AUVs have limitations. If~the steering performance by the rudder is limited, with~the maximum angle of steering $\theta$, the~maximum turning radius should be $R$. For~this reason, the~minimum turning radius should not be less than R, which put constrains on the movements. Most of the traditional path planning methods did not consider the kinematic constraints.
In another situation, a~configurable maximum pitch angle $\gamma$ is set to control the behaviour of the climbing of the AUV. In~case that the necessary pitch angle between two waypoints exceeds this angle, a~helical path is introduced to avoid the generation of pitch set$-$points that might make the vehicle stall.
The Dubins path planning method can handle these two kind of situations caused by the vehicle kinematic constraints~\mbox{\cite{2017Task,2014Path}}.
The Dubins method is mainly based on the following theorem. When the direction of motion is known and the turning radius is the smallest, the~shortest path from the initial vector to the ending vector is composed of a straight line and a turning arc with the smallest radius. AUV can effectively save energy by using the shortest path in the task assignment and path planning.
 {Although the problem of multi$-$task assignment has been extensively studied, none of the literature mentioned above considered AUV's underwater kinematic constraints. This issue has certainly been studied before by some scholars from different perspectives, but~overall, the research is relatively rare. The~AUV underwater task assignment and path planning in the actual environment cannot ignore the kinematic constraints and the actual existence of the obstacle environment. This paper mainly conducts in$-$depth research on this.}

\begin{figure}[H]
	\includegraphics[width=12.5 cm]{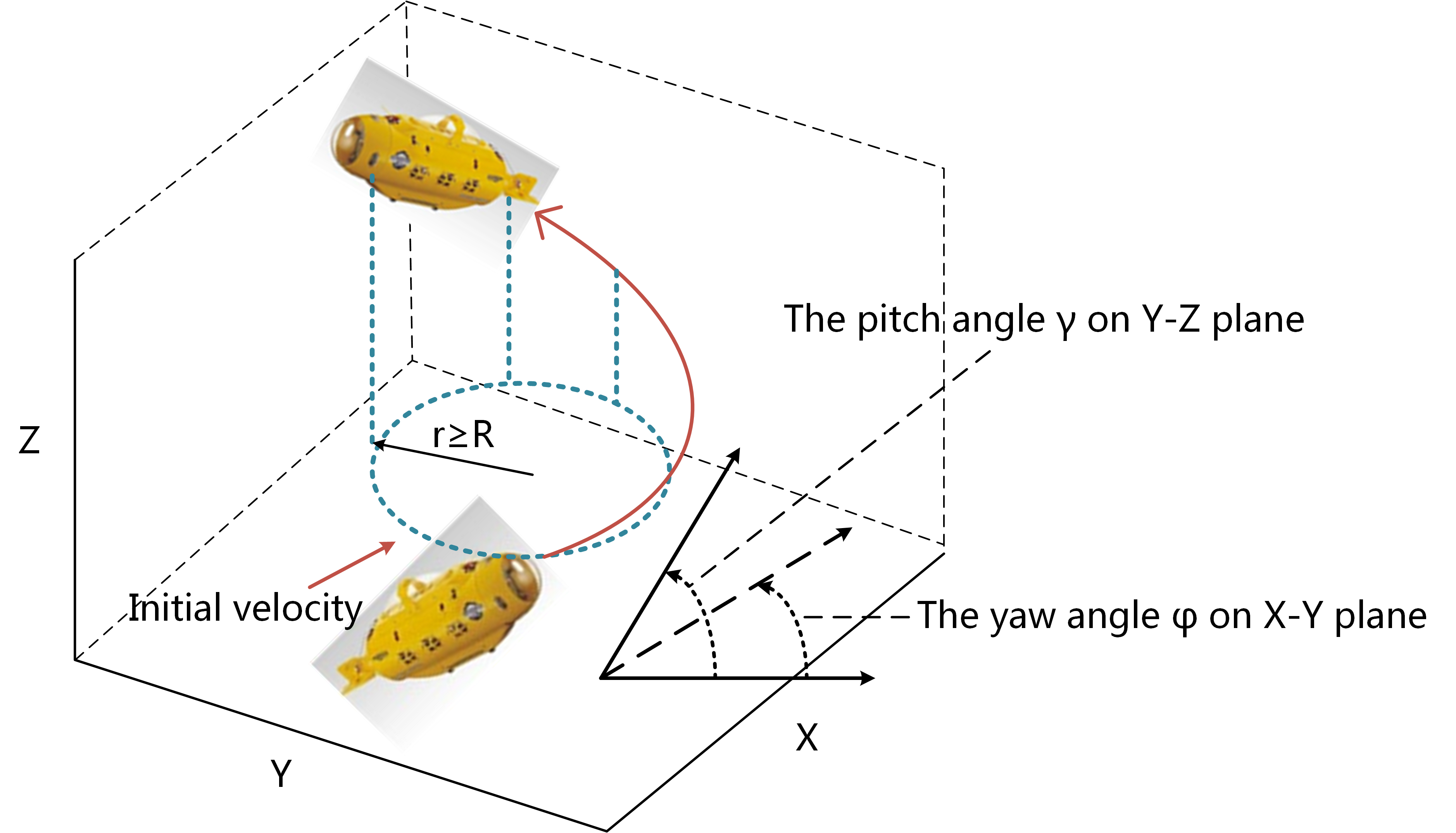}
	\caption{AUV's kinematic constraint in 3D~workspace. \label{kinematic_constraint}}
\end{figure}

As shown in Figure~\ref{kinematic_constraint}, a~typical AUV has 1--2 thrusters at the tail, relying on rudder and fins to control the moving direction. This structure determines that most AUVs are underactuated. When an AUV moves, it turns horizontally and tilts vertically, i.e.,~yaw and pitch. The~yaw and pitch angles are limited due to mechanical structure and fluid dynamics. In~the 3D workspace, pitch angle $\gamma \leqslant \gamma_{max}$, and~yaw angle's derivative or the angular velocity on the X--Y plane $\dot{\varphi}$ is also limited, which results in the turning radius $r \geqslant R$, where $R$ is the minimum turning radius under the angular velocity $max(\dot{\varphi})$. It should be noted that when the force of the thruster is set very high, the~turning radius will also be large at the maximum rudder angle, while we are discussing the case of normal thrust.
Considering this, task assignment and path planning of multiple AUV systems under kinematic constraints in a two$-$dimensional environment are studied in this paper, taking into account static obstacles and other interference factors. The main innovation is to combine the task assignment of SOM neural networks with Dubins path planning under kinematic constraints, making the method more applicable to real$-$world applications. By~combining the Dubins Path algorithm with the improved self$-$organizing mapping (SOM) neural network algorithm, this paper proposes a new task assignment and path planning algorithm, which effectively solves the multi$-$task assignment problem of the multi$-$AUV system in the actual environment. {The~mathematical model and algorithms are established in MATLAB R2016a and~then tested in Python 2.7.18 on a Raspberry Pi.}

This paper is organized as follows.
The task assignment and kinematic motion planning problems are fomulated in Section \ref{sec2}. The~proposed improved SOM neural network method with Dubins path planning algorithms are introduced in Section \ref{sec3}, where the AUVs' kinematics have multi$-$constraints. In~Section \ref{sec4}, simulation results are shown for the task assignment with Dubins path of several AUVs and targets in 2D and 3D workspace even with obstacles. Comparative time comsuptions are recorded to validate the actual availability for applications. In~Section \ref{sec5}, some conclusive remarks are given with some future works discussed. 

\section{Problem~Formulation}\label{sec2}

The task allocation algorithm of a multi AUV system covers two aspects: task allocation and path planning, which refer to the technology of controlling a group of AUVs based on a certain algorithm, and~each of them reaches the target along the optimized path under kinematic~constraints. 

\begin{figure}[H]
	\includegraphics[width=12.5 cm]{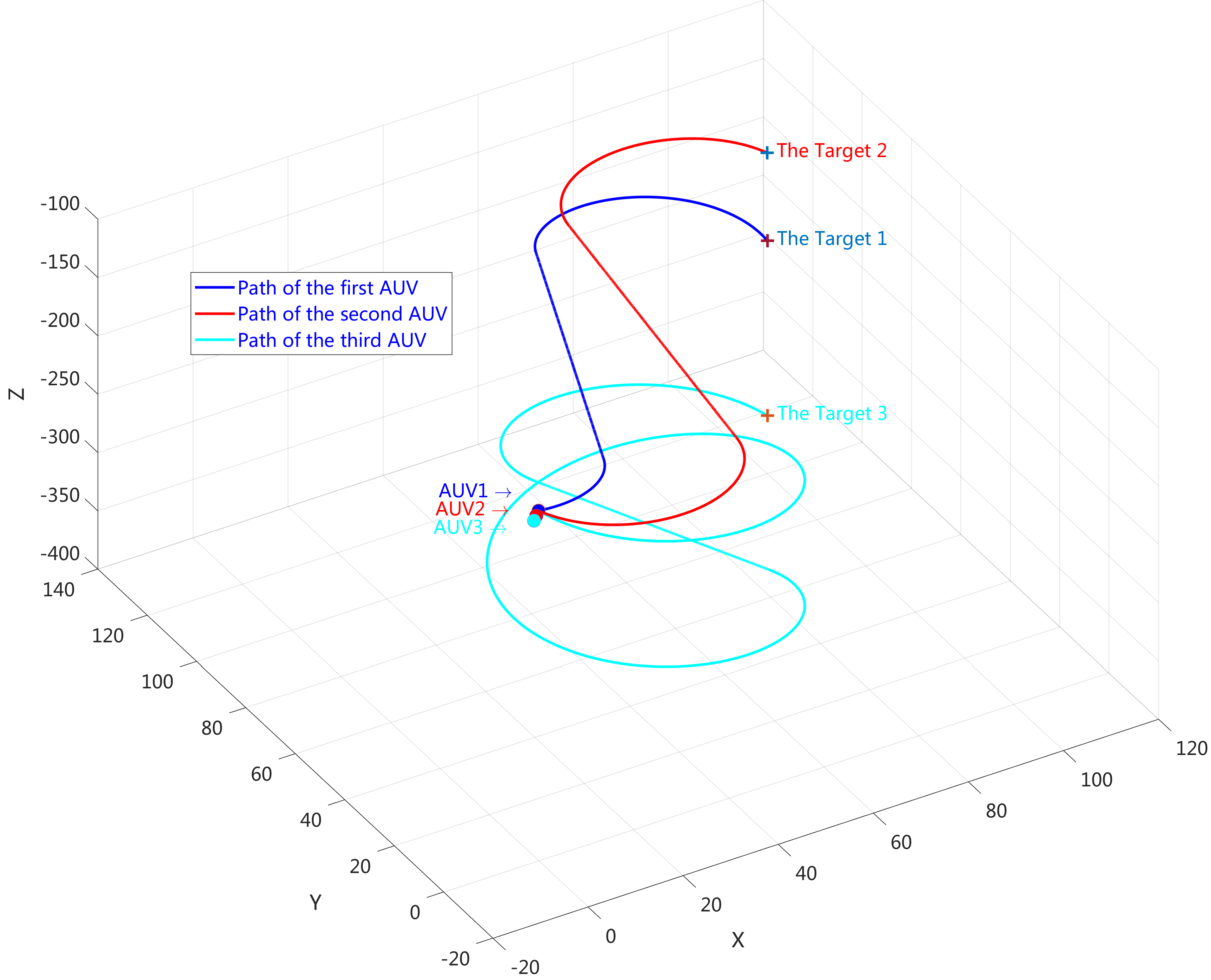}
	\caption{The 
 example of multi$-$AUV task assignment: 3~AUVs and 3~targets. \label{fig1}}
\end{figure}

Task allocation refers to assigning any number of targets to a group of underwater robots, ensuring that all targets are traversed while minimizing the cost of the entire multi AUV system, i.e.,~minimizing the total walking distance. 
Path planning technology refers to a path planning method for a single AUV after global task allocation. 
This article mainly focuses on the task allocation and path planning problem of multi AUV systems under kinematic constraints in two$-$dimensional and three$-$dimensional environments, and~considers interference factors such as obstacles. A~3D visual example is provided in Figure~\ref{fig1}.

Relying solely on the principle of proximity between target point coordinates and AUV coordinates to allocate tasks may sometimes result in one AUV being assigned too many tasks, making it unable to fully reach all assigned targets before running out of power, while other AUVs may not be able to achieve the task objectives.
The initially assigned tasks cannot be completed in the actual planned path due to obstacles. In~order to ensure the maximum utilization of energy carried by each AUV and avoid the AUV stopping due to insufficient energy during its movement towards the target, load balancing of the AUV should be~considered.

Assuming that AUVs have the same underwater cruising capability, carrying the same energy, the~maximum number of each AUV should be decided firstly. Note that we do not consider the energy consumption difference caused by the turning radius for now, only the distance traveled. Firstly, calculate $N=N_T / N_R$, where $N_T$ is the number of target points and $N_R$ is the number of AUVs, then set the upper limit of the task load to
\begin{linenomath}
\begin{equation}
	N_{M A X}= \begin{cases}N, & N \in N^{+} \\ \lfloor N\rfloor+1, & N \notin N^{+}\end{cases}
\end{equation}
\end{linenomath}

When the maximum number of tasks is exceeded the certain number for an AUV, the~task will not be assigned to it again, and~the suboptimal AUV will be selected to execute the task. Considering the load balancing of AUV systems into the algorithm is an important improvement, making them more suitable for the actual situation of multi AUV and multi task allocation. Meanwhile, we consider this problem in reverse and sequentially select the optimal target point for the AUV until the number of tasks for the AUV approaches $N_{M A X}$. At~this point, we then calculate the energy loss of a single AUV to ultimately complete the task~allocation.

After the optimal task assignment, path planning must be implemented with consideration of the kinematic constraints. Assuming that the AUV's initial and final poses are denoted by $P_g\left(x_g, y_g, z_g, \varphi_g, \gamma_g\right)$ and $P_g\left(x_g, y_g, z_g, \varphi_g, \gamma_g\right)$, respectively, where $\varphi_i, \gamma_i$ represent the heading and pitch angles, respectively \cite{yu_wang_real-time_2015,vana_minimal_2020}.
In the Cartesian coordinate frame $(x,y,z)$, the~path connecting the two poses is provided by trajectory smoothing, which can be written as
\begin{linenomath}
\begin{equation}
\left\{\begin{array}{l}
	\frac{d x}{d s}=\cos \varphi(s) \cos \gamma(s) \\
	\frac{d y}{d s}=\sin \varphi(s) \cos \gamma(s) \\
	\frac{d z}{d s}=\sin \gamma(s) \\
	\frac{d \varphi}{d s}=\mu_1 \\
	\frac{d \gamma}{d s}=\mu_2
\end{array}\right.
\end{equation}
\end{linenomath}
where \(s\) represents the curvilinear abscissa along the path, $\mu_1$ and  $\mu_2$ represent the control inputs. The~curvature radius $R(s)$ is calculated as follows:
\begin{linenomath}
\begin{equation}
R(s)=\frac{1}{\sqrt{\mu_1^2(s) \cos ^2 \gamma(s)+\mu_2^2(s)}}
\end{equation}
\end{linenomath}

As there are certain relationships between geometric constraints and the vehicle actual constraints, control inputs $(\mu_1, \mu_2)$ should be determined to satisfy the these geometric constraints: (1) the curvature radius constraint: $|R| \geqslant R_{min}$; (2) the pitch angle constraint: $\gamma_{min} \leqslant \gamma \leqslant \gamma_{max}$. Essentially, finding the shortest path is an optimal control problem. For~a group of AUVs, we are trying to find the optimal total consumption of all AUVs after completing their tasks described as
\begin{linenomath}
\begin{equation}
	\sum_1^N \min \int_0^{s_{gn}} d s
\end{equation}
\end{linenomath}
where $s_{gn}$ represents the planned trajectory length of the $n$th~AUV. 

In this paper, dubins path is used to handle this issue. In~this way, we can plan executable formation trajectories. It is should be noted that the energy loss and formation task reassignment caused by different servo angles of the AUVs are not currently within the scope of this article.
\section{Task Assignment Algorithm under Kinematic~Constraints}\label{sec3}
In this section, self$-$organizing map (SOM) neural network method with Dubins path generator is used to deal with task allocation and path planning problems for the multi$-$AUV system. The~control stability and accuracy of the system model are compensated by strengthening the optimal global exploration and local exploitation ability.  {At the same time, an~event$-$triggered neural network evolution strategy is developed to decrease the
update frequency of the control signal. The~improved optimization method can realize the controller's online parameter adjustment to meet the AUVs' control requirements in the experimental simulation environment.}

\subsection{Application of {Event$-$Triggered} SOM in Multi$-$AUV Task~Assignment}
Assuming a group of AUVs is distributed within a limited working area, and~a random number of targets are distributed within this area. Each target point requires one AUV to complete a specific task at that point. For~each AUV, its cost is measured by the distance it moves from the starting position coordinate point to the target position coordinate point. The~total cost is defined as the sum of all individual AUV costs. After~all target points have been visited once, the~task is~completed.

	For the sake of simplicity, a~two$-$dimensional plane is used as the workspace, in~which the red points represent the AUV, and the green circles represent the target point. Furthermore, assuming that all AUVs are the same robots with basic navigation, obstacle avoidance, and~position recognition functions. For~an input neuron (target point), the~output neuron competes to become the winning neuron, as~follows: 
\begin{linenomath}
\begin{equation}
		\left[\mathbf{N}_j, \mathbf{N}_l\right]=\min \left\{D_{k j l}, k=1, \ldots, K ; j=1, \ldots, J ; l=1, \ldots, L\right\}
	\end{equation}
\end{linenomath}
where $\left[\mathbf{N}_j, \mathbf{N}_l\right]$ represents the $\mathbf{N}_j$th output neuron that competes to win against the $\mathbf{N}_l$th input neuron in the $k$th iteration. $D_{k j l}$ is a weight related variable according to relative distance, defined as:
\begin{linenomath}
\begin{equation}
		\left.D_{kjl}=\begin{cases}\left|\mathbf{T}_l-\mathbf{R}_{jk}\right|\left(1+V\right), &P_j<S_{\max}\\\infty,&P_j\geq S_{\max}\end{cases}\right.
	\end{equation}
\end{linenomath}
where $S_{\max}$ is the maximal distance that a single AUV can travel, and~\begin{linenomath}
\begin{equation} \label{e-distance}
		\left|\mathbf{T}_l-\mathbf{R}_{j k}\right|=\sqrt{\left(x_l-w_{j k x}\right)^2+\left(y_l-w_{j k y}\right)^2}.
	\end{equation}
\end{linenomath}

Equation~(\ref{e-distance}) provides an expression for finding the Euclidean distance between the target location $\mathbf{T}_l$ and the AUV location $\mathbf{R}_{j k}$; specifically, $\mathbf{T}_l=\left(T_{l x}, T_{l v}\right))$ is the coordinate of the input neuron in the Cartesian coordinate system; $\mathbf{R}_{j k}=\left(w_{j k x}, w_{j k y}\right)(k=1,2, \ldots, \mathrm{K}; j=1,2, \ldots, \mathrm{J})$ is the coordinate of the $k$th neuron in the $j$th output neuron group, which is the position of a specific AUV at a certain moment. The~parameter $V$ is given by (\ref{e-load}), which controls the load balancing between various AUVs. The~load balance function is the core of the SOM algorithm. The~winning neuron in competition is not only the neuron with the smallest Euclidean distance from the input neuron but~also the neuron with the smallest load at that moment. 
\begin{linenomath}
\begin{equation} \label{e-load}
		V=\frac{P_j-\bar{v}}{1+\bar{v}}
	\end{equation}
\end{linenomath}
where $P_j$ is the path length of the $j$th AUV's movement, and~$\bar{v}$ is the average path length of the AUV~team.

\begin{figure}[H]
	\includegraphics[width=14 cm]{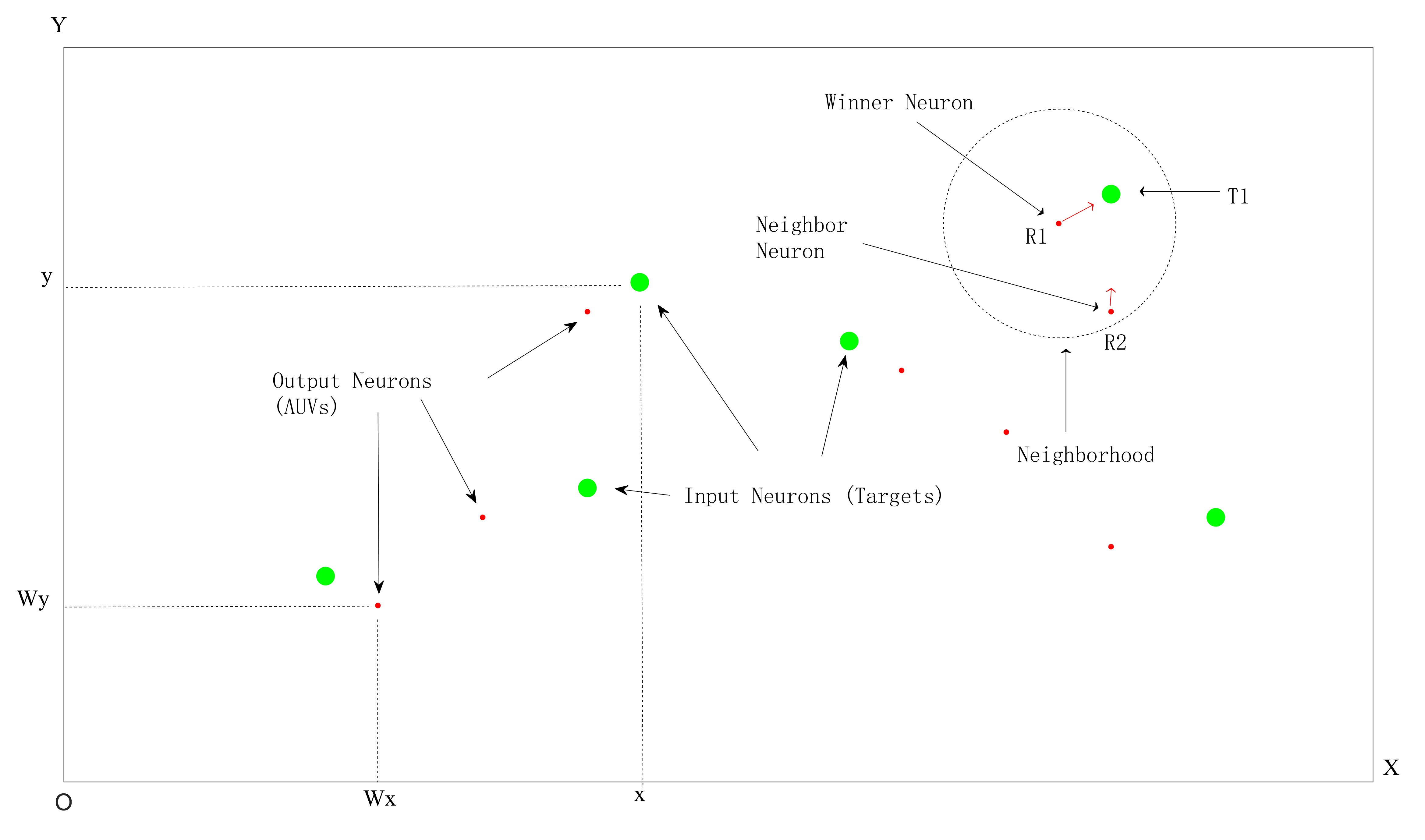}
	\caption{Schematic diagram of neighborhood weight change and load~balance. \label{fig2}}
\end{figure}

After a AUV's representing neuron wins the competition, a~neighborhood function is an important step. The~neighborhood function determines the influence (attraction strength) of the input neuron on the winning neuron and its adjacent neurons. The~impact on the winning neuron should be the greatest. The~impact on neighboring neurons gradually decreases, while neurons outside the neighboring area are not affected. The~magnitude of the impact determines the size of the weight adjustment of neurons in the neighborhood during a certain iteration process. The~process of calculating neighborhood functions and changing weights is shown in Figure~\ref{fig2}. The~red dots represent the position of the AUV at a certain moment, serving as output layer neurons. The green circle represents the target position as the input neuron. The~winning neuron in the figure is $R_1$, which is one closest to the input $T_1$. The~neighborhood function is defined as follows:
\begin{linenomath}
\begin{equation} \label{}
		f\left(d_j, G\right)= \begin{cases}e^{-d_j^2 / G^2(t)}, & d_j<r \\ 0, & \text { other }\end{cases}
	\end{equation}
\end{linenomath}

It is easy to see that $0 \leq f\left(d_j, G\right) \leq 1$, where $d_j=\left\|\mathbf{j}-\mathbf{N}_l\right\|$, representing the distance between the $j$th neuron in the $k$th group and the winning neuron $\mathbf{N}_l$. $r$ is the neighborhood radius. $ G(t)=(1-m)^t G_o$, where $t$ is the number of iterations. When $t$ increases, the~value of the neighborhood function decreases, and~the moving step of the AUV in the neighborhood decreases. $m$ and $G_o$ are constants, through which the motion step size of AUVs in the neighborhood can be adjusted, thereby controlling calculation accuracy and operation~time.

After the winning neuron and its neighborhood are determined, the~winning neuron and adjacent neurons move towards the input neuron, while the other neurons remain stationary. The~update rule is given by (\ref{weight}), where the addition operator represents vector addition. The~termination condition for the operation is provided by $D_{min}$, which can reduce the calculation time. The~modification of weight values depends not only on the initial distance between the winning neuron and its neighboring neurons and the input target point but~also on the neighborhood function and the learning rate of the network $\alpha$.
\begin{linenomath}
\begin{equation} \label{weight}
		\mathbf{R}_{jk}(t+1)=\begin{cases}\mathbf{T}_{l}(t),&D_{kjl}<D_{\mathrm{min}}\\\mathbf{R}_{jk}(t)+\alpha\cdot f(d_{j},G)\cdot\left(\mathbf{T}_{l}(t)-\mathbf{R}_{km}(t)\right),&\text{other}\end{cases}
	\end{equation}
\end{linenomath}

At the same time, if~there exists an obstacle at the position $ob = (x_o,y_o)$, the~obstacle weight is computed based on the Euclidean distance between it and the neural network weights by
\begin{linenomath}
\begin{equation} \label{ob-distance}
		ob_w = min(\left|\mathbf{ob}-\mathbf{R}_{j k}\right|)
	\end{equation}
\end{linenomath}
where $\mathbf{ob}$ is the expansion of $ob$ to a vector of equal length to $\mathbf{R}_{j k}(\mathrm{j}=1,2, \ldots, \mathrm{J};$ \linebreak\mbox{$\mathrm{k}=1,2, \ldots, \mathrm{K})$}. By~comparing $ob_w$ with the defined safety distance $d_safety$, the~impact of obstacles can be~determined.

\begin{figure}[H]
	\includegraphics[width=12 cm]{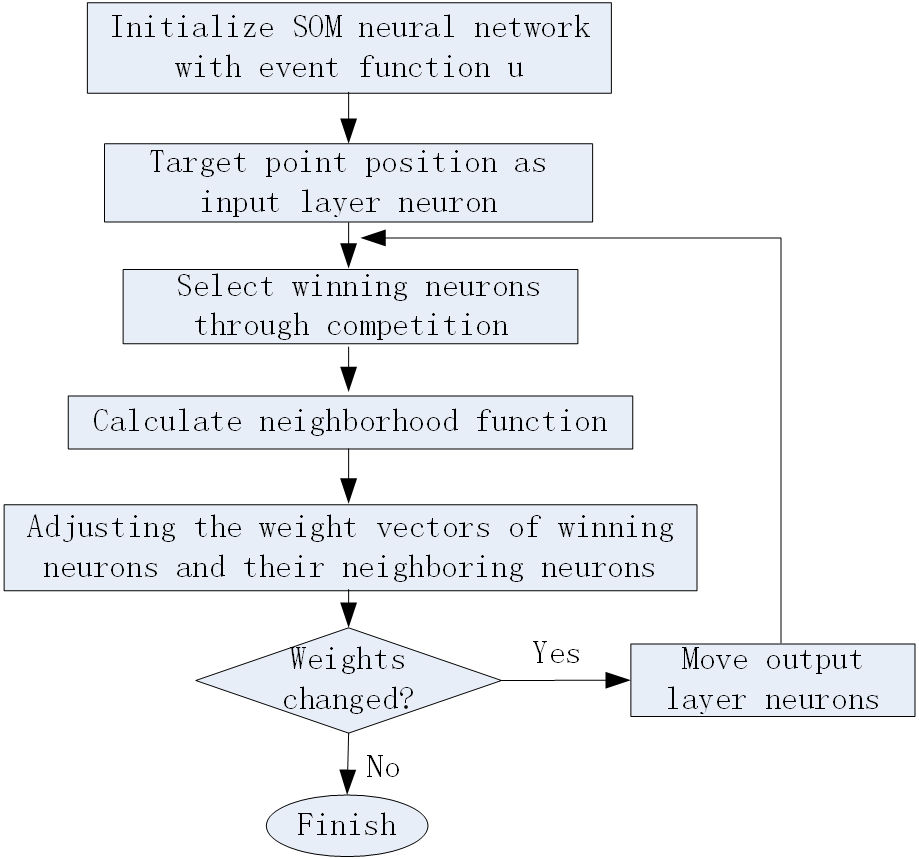}
	\caption{The event$-$triggered SOM algorithm flow~chart. \label{fig3}}
\end{figure}

The flowchart of the algorithm is shown in Figure~\ref{fig3}. After~initializing the SOM neural network, the~positions of the target points are sequentially input into the network. For~a given input target point during the iteration process, it can be summarized as a three$-$step calculation process. The~first step is to select the winning neuron; the second step is to determine the neighborhood of the winning neuron; the third step is to correct the weight vectors of the winning neuron and its neighboring neurons. 
 {Event triggering occurs in three situations, namely obstacles on the path, load exceeding the upper limit, or~inability to plan the Dubins path from the AUV to the pre$-$allocated target point. The~event triggered control functions $u$ is given by (\ref{event1}), where $u_1$ is the workload related function by (\ref{wk}) and  $u_2$ is the obstacles related function by (\ref{ob}). The~event triggered re$-$assignment will be implemented in the SOM algorithm flow chart if $u = 0$.
After all target points are reached by one specified AUV, task allocation is completed.}
\begin{linenomath}
\begin{equation} \label{event1}
		u = u_1 \&u_2
	\end{equation}
\end{linenomath}
\begin{linenomath}
\begin{equation} \label{wk}
		u_1 = \begin{cases} 0, & D_{kjl} = \infty \\
			   1, & others	\end{cases}
	\end{equation}
\end{linenomath}
\begin{linenomath}
\begin{equation} \label{ob}
			u_2 = \begin{cases} 0, & ob_w > d_{safety} \\
			 					1,  & ob_w \leq d_{safety} 
					\end{cases}
	\end{equation}
\end{linenomath}

\vspace{-6pt}
\unskip

\unskip

\subsection{Dubins Path Planning Algorithm for AUV Kinematic~Constrains}
 {Unlike underwater robots such as ROVs, most torpedo like AUVs and underwater gliders are underactuated with tail thrusters to only moving forward, steering depend on fins and rudders.} It is difficult to achieve proper path planning. Fortunately, Dubin's Car was introduced into the literature by Lester Dubins, a~famous mathematician and statistician, in~a paper published in 1957. The~cars essentially have only 3 controls: “turn left at maximum”, “turn right at maximum”, and~“go straight”. All the paths traced out by the Dubin's car are combinations of these three controls. Dubin's car is very similar to AUVs' kinematics.  {Assuming that the AUV has an initial velocity and dynamic characteristics as mentioned above, the~path can be planned based on the Dubins car method from 2D to 3D workspace. Precise motion control to dynamic control problem is also very useful, which is another quite challenging field being studied. For~example, command filter with adaptive control was proposed to solve the model uncertainties and input saturation issues~\cite{liu_adaptive_2023}. These theories will be involved in our future work. To~minimize the traveling time and energy consumption of AUVs, path planning based on kinematic is necessary. Here, we adopt the Dubins method to deal with the three$-$dimensional motion planning problem.}

Literally, the~Dubins curve is the shortest path connecting two points with initial directions, while satisfying curvature constraints and specified tangent lines (entry direction) at the beginning and end, and~limiting the target to only turn forward. As~described in prior works, the~Dubins curve can be represented as a combination of three basic movements, as~listed in Table~\ref{tab1}. The~Dubins curve provides a sufficient set of paths, which includes the optimal path. The~shortest path is only selected from 6 curves in the set of {LRL LSL LSR RLR RSR RSL}.

\begin{table}[H] 
	\caption{Table of concise classification of the Dubins~path.\label{tab1}}
	\newcolumntype{C}{>{\centering\arraybackslash}X}
	\begin{tabularx}{\textwidth}{CCC}
	\toprule
	\textbf{Symbol}	& \textbf{Meaning}	& \textbf{Direction}\\
	\midrule
	L		& Turn left			& Counterclockwise\\
	R		& Turn right			& Clockwise\\
	S		& Go straight			& Forward\\
	C       & Circular arc         & Na \\
	CCC		& 3 arcs				& {LRL RLR} \\
	CSC	 	& 2 arcs and 1 line segment & {LSL RSR LSR RSL} \\
	\bottomrule
	\end{tabularx}
\end{table}

\begin{figure}[H]
	\includegraphics[width=11.5 cm]{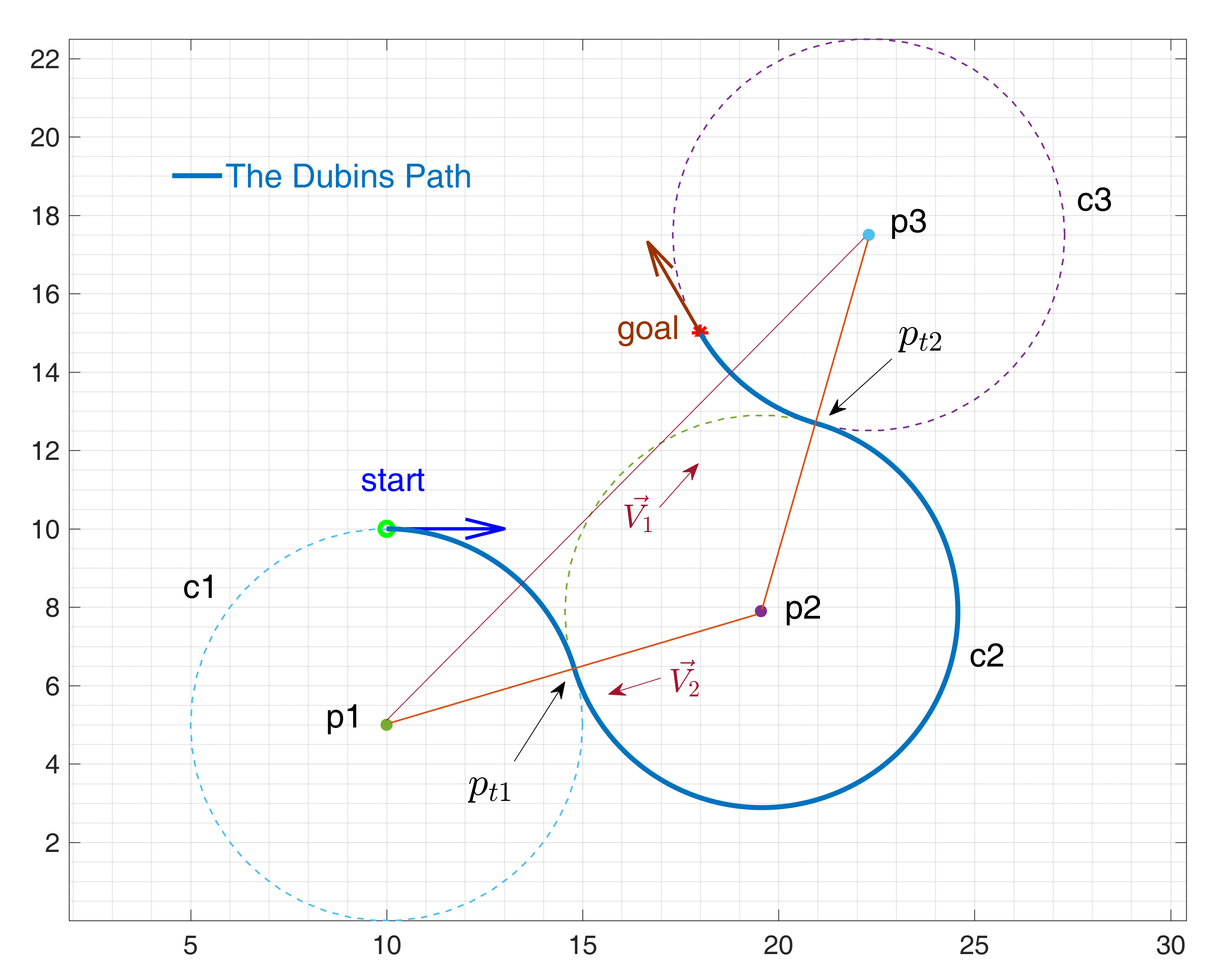}
	\caption{Computing a RLR Dubins~trajectory.} \label{fig_ccc}
\end{figure}

Dubins curves in above table have been discussed in several papers such as~\cite{2014Path,yu_wang_real-time_2015}, and~related calculation methods are provided in former research, which will not be repeated in this article. Two of the Dubin's shortest paths do not use the tangent lines. These are the RLR and LRL trajectories, which consist of three tangential, minimum radius turning circles  {\cite{giese_comprehensive_nodate, shkel_classication_2001}. These trajectories may be used very often in the swarm situation, where a lot of AUVs densely distributed in a certain 3D workspace. If~the distance between the agent and the target's turning circles is less than 4 times the minimum turning radius $r_{min}$, then a CCC curve is valid. For~CCC trajectories as an example, we must calculate the location of the third circle, as~well as its tangent points to the circles. To~be more concrete, the~RLR case is shown in Figure~\ref{fig_ccc}. }

Consider the minimum$-$radius turning circle to the right of start location to be the circle c1, and~the minimum$-$radius turning circle to the right of the goal location to be the circle c3. The~task is now to compute c2, a~circle tangent to both c1 and c2 plus the points $p_{t1}$ and $p_{t2}$ which are respective points of intersection between the 3 circles. Let \textbf{p1}, \textbf{p2}, \textbf{p3} be the centers of circle c1, c2, and~c3,respectively. The~triangle is formed using these points. Because~c2 is tangent to both c1 and c2, the~lengths of all three sides are~known.

Segments $\overline{\textbf{p1} \textbf{p2}}$ and $\overline{\textbf{p3} \textbf{p2}}$ have length $2r_{min}$, and segment $\overline{\textbf{p1} \textbf{p3}}$ has length $d$. We are interested in the angle $\theta=\angle \textbf{p3}\textbf{p1}\textbf{p2}$ because that is the angle that the line between c1 and c3 ($\vec{V}_{1}$) must rotate to face the center of c2, which will allow us to calculate \textbf{p2} by $\theta=cos^{-1}\left(\frac{d}{4r_{min}}\right)$, where $\theta$ in a RLR trajectory represents the amount of rotation that vector $\vec{V}_{1} = \textbf{p3} - \textbf{p1}$ must rotate to point at the center of c2; $d = \sqrt{\left(x_{3}-x_{1}\right)^{2}+\left(y_{3}-y_{1}\right)^{2}}$. However, $\theta$'s value is only valid if $\vec{V}_{1}$ is the same direction as the positive x$-$axis. Otherwise, the~atan2 function will be needed to rotate $\vec{V}_{1}$. For~a LRL trajectory, we want to add $\theta$ to this value, but~for an RLR trajectory, we want to subtract $\theta$ from it to obtain a circle at the right$-$side of c1. Note that we consider counter$-$clockwise turns to be positive. Now that theta represents the absolute amount of rotation, we can compute the c2 center point
\begin{linenomath}
\begin{equation} \label{compute_d1}
		\textbf{p2}=(x_1+2r_{min}cos\left(\theta\right),y_1+2r_{min}sin\left(\theta\right))
	\end{equation}
\end{linenomath}
after which the tangent points $p_{t1}$ and $p_{t2}$ becomes easy to be computed. Defining vectors from the \textbf{p2} to \textbf{p1} and \textbf{p3}, and~walking down them a distance of $r_{min}$, we obtain the vector $\vec{V_2}$ from \textbf{p2} to \textbf{p1}. Next, change the vector's magnitude to $r_{min}$ by normalizing it and multiplying by $r_{min}$
\begin{linenomath}
\begin{equation} \label{compute_d2}
		\vec{V_2}=\frac{\vec{V_2}}{\|V_2\|}*r_{min}
	\end{equation}
\end{linenomath} 
where $\vec{V_2} = \textbf{p1} - \textbf{p2}$. Next, compute $p_{t1}$ using the new $\vec{V_2}$ by $p_{t1}= \textbf{p2} + \vec{V_2}$. Then, $p_{t2}$ will be computed following a similar~procedure.

\vspace{-3pt}

As the tangent points are obtained, arc lengths and duration as before can be computed to finish things off. Then, we obtain a Dubins path composed of 3 curves known as the CCC trajectory.
To extend the 2D Dubins curves to 3D space using the linear interpolation method, the~3D tour sequences are first projected on to the 2D $[X,Y]$ plane in a global coordinate system. Taking a starting point $P_0(X_0, \Phi_{0}]$ and an ending point $P_1(X_1, \Phi_1)$ in the 3D $[X,Y,Z]$ space and project them on to the 2D plane. Then, the starting and ending points become 2D parameters $[(x_0,y_0),\phi_0]$ and $[(x_1,y_1),\phi_1]$. The~2D Dubins curve is designed as described in the above content, and~the lengths of the arcs and line segment are calculated. Let $\mathcal{L}_{0,x}$
and $\mathcal{L}_{x,1}$ denote the lengths along the 2D Dubins curve from $(x_0,y_0)$ to (x, y) and from $(x,y)$ to $(x_1,y_1)$, respectively. The~linear interpolation adds the z coordinate in the 3D space by
\begin{linenomath}
\begin{equation} \label{compute_3D}
		z=z_0+\frac{\mathcal{L}_{0,x}}{\mathcal{L}_{0,1}}(z_1-z_0)
	\end{equation}
\end{linenomath}
where $z_0$ and $z_1$ are the Z coordinates of the starting and ending points. A detailed procedure has been introduced in existing papers such as~\cite{2017Task,vana_minimal_2020} and will not be repeated here. If~obstacles appear in the path, we must re$-$assign the~tasks. 

After obtaining the dubins path, we modified the task allocation algorithm as following Figure~\ref{algorithm2}. With~kinematic constraints considered, the~computing process increases the base on algorithm 1. Due to the limited energy carried by autonomous underwater robots themselves, the~rational and effective utilization of energy has become a key constraint for multi$-$AUV systems to complete multiple tasks. Therefore, the~load balancing of AUVs is considered in the algorithm. The~paper assumes that AUVs carry the same energy, that is, within~the same working range, the~distances traveled by AUVs exhausting their own energy are the same. Load balancing is determined by the traveling distance limit. At~the same time, it is determined whether the distance between the winning AUV and the target task point meets the conditions for forming the Dubins Path. If~not, it is set to infinity. Then the input neuron data could be modified in Algorithm 1 to reassign tasks to appropriate~AUVs.

\begin{figure}[H]
	\includegraphics[width=12.5 cm]{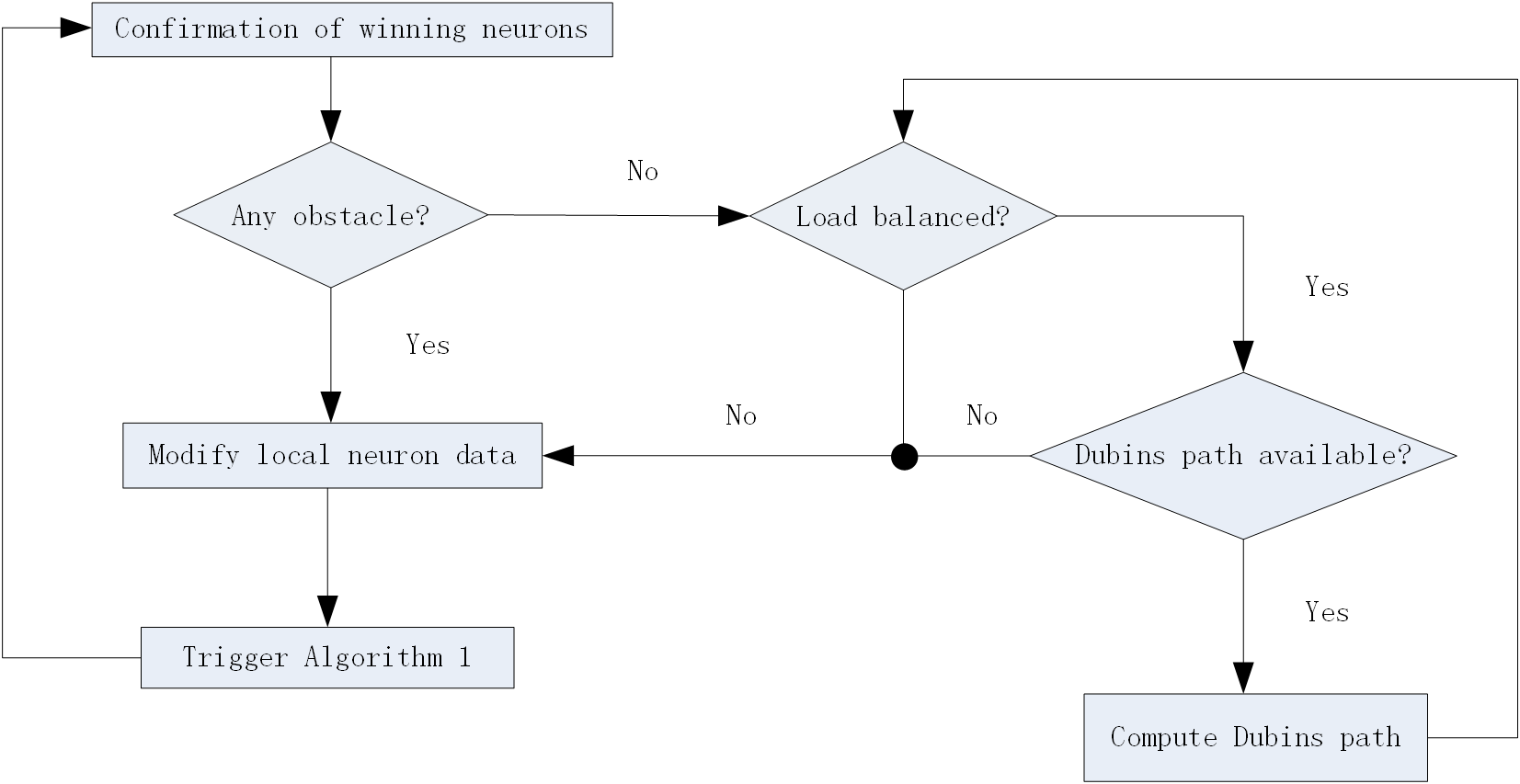}
	\caption{Flow chart of task assignment algorithm with Dubins~path. \label{algorithm2}}
\end{figure}
\unskip

\section{Simulation~Research}\label{sec4}

\begin{figure}[H]
	\begin{adjustwidth}{-\extralength}{0cm}
	\centering
	\includegraphics[width=17cm]{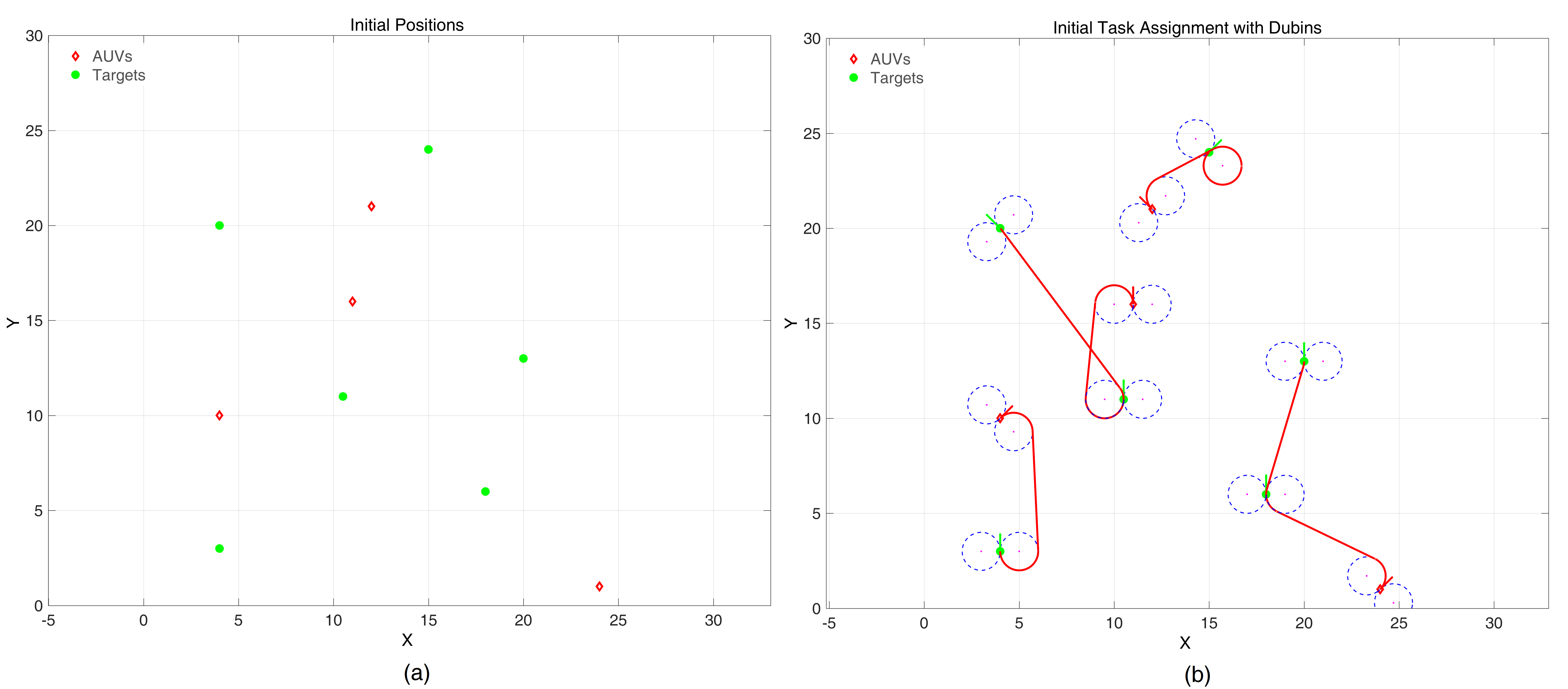}
	\end{adjustwidth}
	\caption{Task assignment in an initial obstacle free and load balance free~environment. (a) initial positions; (b) assignment result. \label{initial_dubins}}
	\end{figure}

Simulations were set up with multiple underwater targets deployed randomly in a work space of
$[30 \times 30]$ units, where the minimum turning radius $r_{min}$ of each AUV are the same and equal to 1 unit. Green dots represent targets, and~red diamonds represent AUVs. In~the workspace of a multi AUV system, there are four AUVs that need to access 6 randomly distributed targets. After~the task is assigned by the SOM neural network, each AUV can reach its nearest target point along the optimal path, as~shown in Figure~\ref{initial_dubins}. As~there is no obstacle or load balance problem in this workspace, AUVs can accurately reach various target positions based on proposed~algorithm. 

\vspace{-6pt}

In actual operation, there are often obstacles that affect the results of task allocation. We keep the positions of AUVs and target points unchanged while adding an obstacle area to the workspace, which may also affect the load~balancing.

\begin{figure}[H]
	\begin{adjustwidth}{-\extralength}{0cm}
	\centering
	\includegraphics[width=17cm]{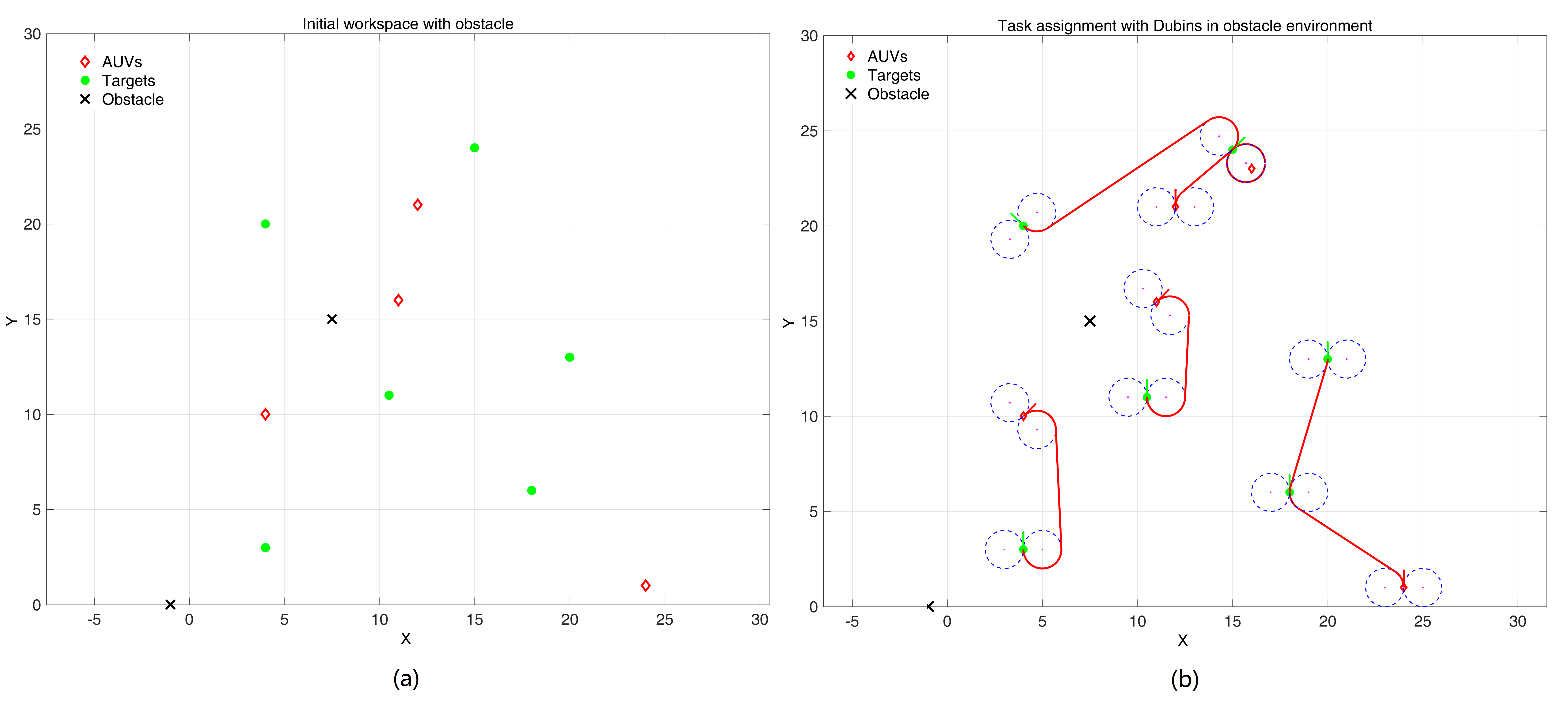}
	\end{adjustwidth}
	\caption{Task assignment and path planning in an obstacle~environment. (a) initial positions of AUVs, targets and obstacles; (b) simulaiton result. \label{ob_dubins}}
\end{figure}

As Figure~\ref{ob_dubins} illustrates, comparing Figure~\ref{ob_dubins}b to Figure~\ref{initial_dubins}b, it can be found that when the route is blocked by the obstacle or the path cannot be planned, the~current winning neuron is not selected, and~the suboptimal neuron is used for route planning until the path planning is successful after the task re$-$assignment. All target points are also accessed by the closest distance. The~simulation results prove that the algorithm is effective with~obstacles. 

In more cases, besides~obstacles, there is also the issue of load balancing for AUVs. AUVs navigate in water and carry limited energy, which means the total range that a single AUV can navigate is limited. The~algorithm proposed in this article can solve these problems within a certain range. The~following is a simulation~explanation. 

\begin{figure}[H]
	\begin{adjustwidth}{-\extralength}{0cm}
	\centering
	\includegraphics[width=17cm]{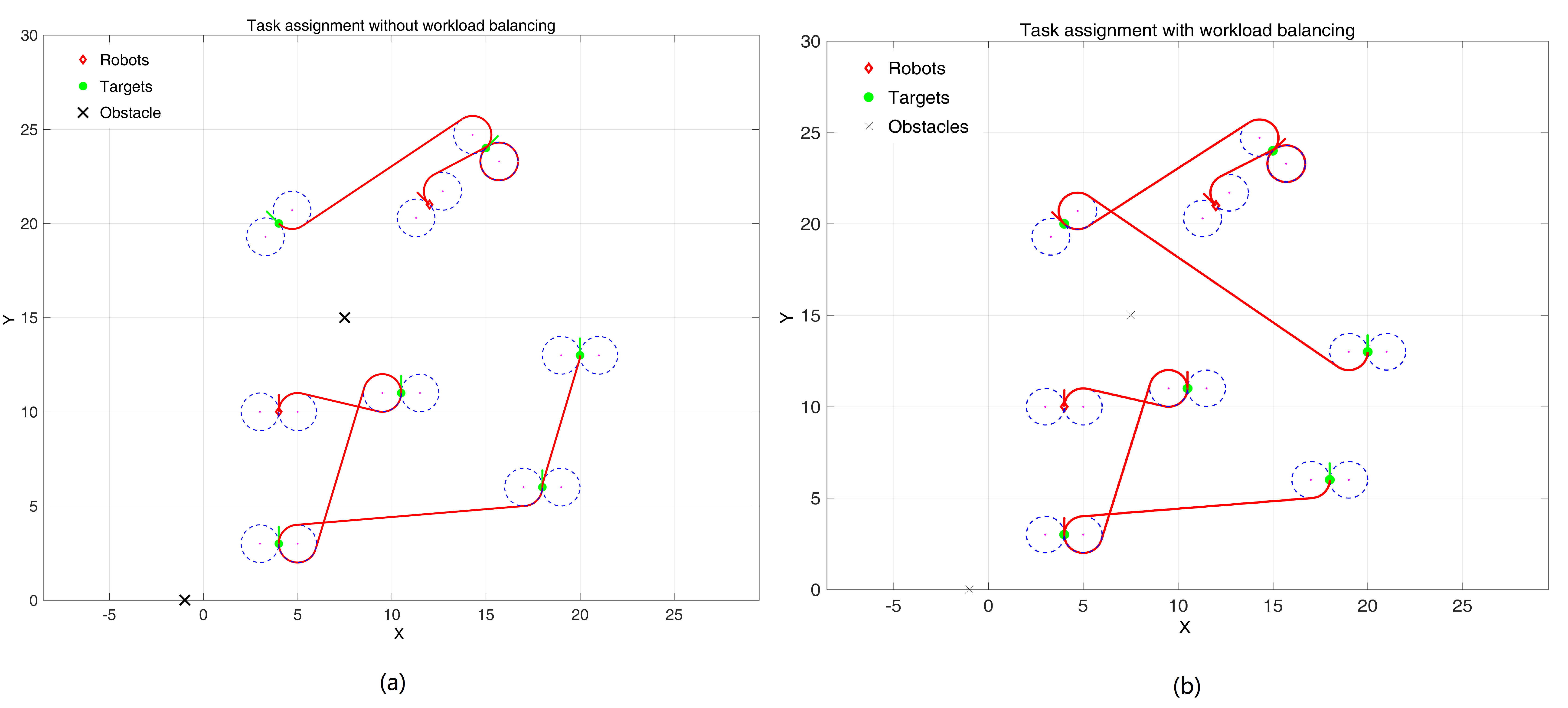}
	\end{adjustwidth}
	\caption{Task assignment in an obstacle environment with workload~balancing. (a) initial state; (b) assignment result. \label{wb_dubins}}
\end{figure}

As shown in Figure~\ref{wb_dubins}, in~the workspace of a multi$-$AUV task assignment scenario, there are 2 AUVs that need to access 6 randomly distributed targets. 
Figure~\ref{wb_dubins}a shows task allocation and path planning without workload balance settings, where one AUV is responsible for four tasks and the other is responsible for two tasks. Figure~\ref{wb_dubins}b shows task allocation and path planning with a setting of number of tasks as 3. At~this time, both AUVs are responsible for completing 3 tasks, and~their energy consumption is relatively balanced. In~the algorithm, the~upper limit of task responsibility is set by default to the number of tasks divided by the number of robots, rounded up by 1. At~the same time, the~distance traveled is also recorded, and~the maximum walking path upper limit is provided. When the upper limit is exceeded after the execution of this task, the~task is not assigned to the winning robot. A~series of tests are implemented as illustrated in Table~\ref{tab_2}. The~path length is using ``Unit'' as a measurement unit as mentioned~above.

\begin{table}[H]
\caption{Task assignment and path planning results with and without workload~balancing.\label{tab_2}}
	\begin{adjustwidth}{-\extralength}{0cm}
		\newcolumntype{C}{>{\centering\arraybackslash}X}
		\begin{tabularx}{\fulllength}{CCCCCCC}
			\toprule
			\textbf{AUV Number ($\mathbold{n}$)}	& \textbf{Target Number} & \textbf{Whether Load Balanced} & \textbf{Path Length (Total)} & \textbf{Path Length (Max)} & \textbf{Tic Toc on PC (ms)} & \textbf{Clock on Pi (ms)}\\
			\midrule
\multirow{4}{*}{2}	& 4			& No			& 8.2 & 4.6	 & 23   &  nul\\
			  	    & 6			& No			& 10.5 & 6.4 	& 25 &  3550  \\
					& 4			& Yes			& 8.2 & 4.6 	& 33  &  nul \\
			  	    & 6			& Yes			& 11.1 & 5.6 	& 52  &  5012 \\
                   \midrule
\multirow{4}{*}{4}    & 6			& No			& 10.6	& 3.0	& 30    &  nul\\
			  	         & 8			& No			& 15.2	& 5.7	& 32 & nul\\
			             & 6			& Yes			& 11.5	& 2.2	& 51 &  nul\\
						 & 8			& Yes			& 16.1	& 4.1	& 55 &   nul\\
                   \midrule
\multirow{4}{*}{6}    & 8			& No			& 15.9	& 3.5	& 46    & 5276\\
			  	         & 10			& No			& 20.2	& 5.6	& 49 & nul\\
			            & 8				 & Yes			& 16.7	& 2.8	& 60 & 7189\\
						& 10 			& Yes			& 23.8	& 4.3	& 67 & nul\\
			\bottomrule
		\end{tabularx}
	\end{adjustwidth}
\end{table}

{As can be seen in the table above, the~distance traveled by AUVs varies depending on whether load balancing is enabled or not. The~algorithms were firstly implemented with MATLAB R2016a software on the hardware platform of HP Computer with Intel Core i7 9700 CPU and 16G DDR4 2400 memory. Simulation run time consumption was recorded using the Tic and Toc function. When there is no load balancing, the~AUV obtains the minimum total distance based on the operation results of the SOM neural network, but~the traveling length of a certain AUV may deviate from the average value. In~the case of 4~AUVs for 8 targets, we calculate the mean square deviation of the distance based on the recorded running distance of each AUV, and~the calculated formula is $\tilde{L}=\sqrt{\frac{\sum_{i=1}^n(L_i-\overline{L})^2}{n-1}}$. Then, the standard deviation of the sample without load balancing can be obtained as 1.812; the standard deviation of the sample with load balancing can be obtained as 0.133. If~there is no load balance, the~load of a certain AUV may exceed the mean value significantly, which is not feasible in practical applications. Thus, we obtain the multi$-$task allocation results based on Dubins path planning under load balancing. Under~load balancing, due to the fact that the pairing selection between certain AUVs and the target points may not be optimal in distance, the~total path length will increase and the computation time will also increase, which are all within an acceptable range on the level of tens of milliseconds. The~simulation results show that the algorithm considering workload balance is effective.
}

{Meanwhile, we implemented the algorithm through Python programs, with~package NumNy. On~the PC, the~simulation time consumption with Python is approximately 5--10~times that of Matlab. In~order to get closer to practical applications, we run Python simulations on the embedded device, an~official Raspberry Pi 4B with 4G memory. The~running times are shown in the table, which basically take a few seconds to complete the core algorithm, without~drawing the graphics. Considering the rapid development of ARM computing power, it can be said feasible to run this algorithm on embedded systems. It should be noted that running on the Pi is only an experiment for the core program, not in an actual operation. The~field applications may require an omniscient perspective and a known global map (or SLAM), which may involve various additional hardware.}

\begin{figure}[H]
	\begin{adjustwidth}{-\extralength}{0cm}
	\centering
	\includegraphics[width=18cm]{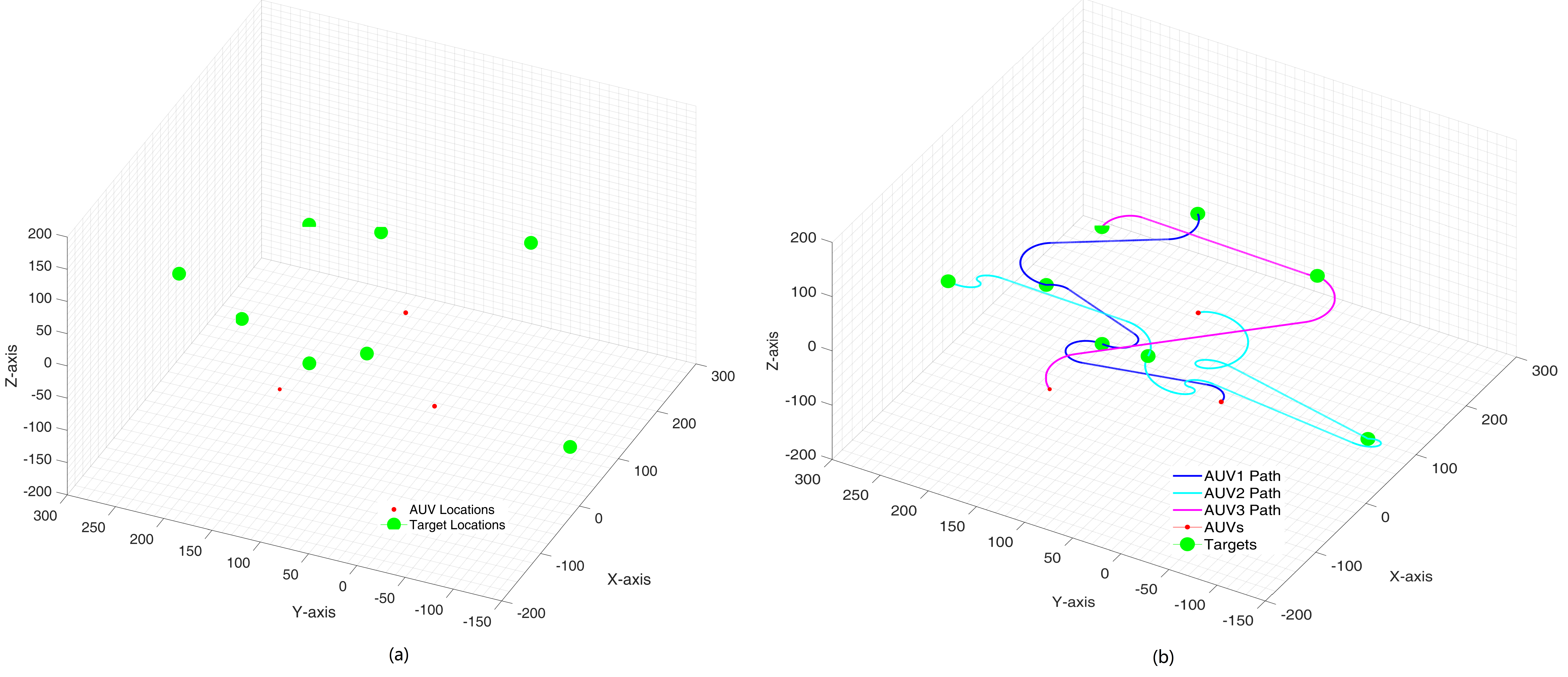}
	\end{adjustwidth}
	\caption{Task assignment in 3D environment with workload~balancing. (a) targets and AUVs initial positions; (b) simulation result visualization. \label{3DA_dubins}}
\end{figure}

Based on the method discussed in Section~\ref{sec2}, we extend the application to three$-$dimensional workspace. Task assigment and Dubins paths in 3D space also face obstacles and load balancing problems. Due to paper length limitations, we will provide an example of simulation here, without~considering dynamic disturbances caused by water flow. Simulations were set up with multiple underwater targets deployed randomly in a 3D workspace, as~shown in Figure~\ref{3DA_dubins}a. Note that in a 3D environment, the pitch angle $\gamma$ is taken into algorithm and restricted to $[-15*PI/180,15*PI/180]$. Task allocation results and trajectories are illustrated in Figure~\ref{3DA_dubins}b for 3 AUVs and 8 targets. We have simulated 30 Monte Carlo scenarios with AUVs and targets, and~computed the average length of 3D trajectories. The~standard deviation value $\tilde{L}$ is relatively small since the proposed algorithms can achieve progressive optimization using the load balancing scheme. The~comparison results related to energy balance are shown in Table~\ref{tab_3}. {We also implemented some simulations on PC (MATLAB) and Pi (Python). The~average running time within an acceptable range has been recorded in the~table.}

\unskip



\begin{table}[H]
	\caption{{Task assignment and path planning results in 3D workspace with and without workload~balancing.}\label{tab_3}}
		\begin{adjustwidth}{-\extralength}{0cm}
			\newcolumntype{C}{>{\centering\arraybackslash}X}
			\begin{tabularx}{\fulllength}{CCCCCC}
				\toprule
				\textbf{AUV Number ($\mathbold{n}$)}	& \textbf{Target Number} & \textbf{Average Path Length} & \textbf{Standard Deviation ($\mathbold{\tilde L}$)} & \textbf{Tic Toc on PC (ms)} & \textbf{Clock on Pi (ms)}\\
				\midrule
	\multirow{3}{*}{3}    
							& 8			& 103.5	& 1.005 & 78 & 8863	\\
							 & 10		& 131.7	& 0.516 & 81 & nul\\
							 & 15		& 206.0	& 0.232 & 101 & nul \\
					   \midrule
	\multirow{3}{*}{5}   
						& 10			& 96.1	& 0.710 & 82 &	9012\\
							& 15		& 129.6	& 0.425 & 89 &	nul\\
							& 20 		& 260.5	& 0.951 & 103 &	nul\\
				\bottomrule
			\end{tabularx}
		\end{adjustwidth}
	\end{table}

\section{Conclusion and Future~Work}\label{sec5}

This paper has studied the task assignment and path planning problem for multi$-$AUV systems with kinematic character considered. The~improved SOM neural network method based on workload balance and neighborhood function is adopted to slove the strategy$-$level issues for task allocation. Meanwhile, based on feasible kinematic path planning, combined with the Dubins method for path planning and task reassignment, task allocation and path planning under load balancing are finally achieved.  {The 2D and 3D Dubins curves are designed with a set of possible headings and with nonholonomic motion constraints. It is demonstrated that the proposed method has been proven to achieve feasible load balancing for task assignment in obstacle environments. In~future work, firstly, we will rewrite the algorithm in C++, hoping to achieve faster speed on embedded systems. Secondly, we will study more practical application and controllers related to dynamic control methods~\cite{WOS:000792022100007}, such as Dubins trajectory tracking problem in water flow workspace of multi$-$AUV systems, as~well as inter group collision avoidance of AUVs on the Dubins paths, and~the moving obstacle avoidance problem.}





\vspace{6pt} 




\authorcontributions{ }

\funding{This research received no external funding.}

\institutionalreview{Not applicable.}

\informedconsent{Not applicable.}

\dataavailability{\href{https://github.com/ayawaya2014/data-for-sensors-paper-202406}{3D Task Allocation and Dubins Trajectories Data of Figure~10}}



\acknowledgments{The authors would like to thank all the Editors and Reviewers who have given their valuable advice to this~paper.}

\conflictsofinterest{The authors declare no conflicts of~interest.} 

\begin{adjustwidth}{-\extralength}{0cm}

\reftitle{References}

\PublishersNote{}
\end{adjustwidth}
\end{document}